\documentclass{article}

\usepackage[final]{neurips_2023}

\usepackage[utf8]{inputenc} 
\usepackage[T1]{fontenc}    
\usepackage{hyperref}       
\usepackage{url}            
\usepackage{booktabs}       
\usepackage{amsfonts}       
\usepackage{nicefrac}       
\usepackage{microtype}      
\usepackage{xcolor}         
\usepackage{amsmath}
\usepackage{graphicx}
\usepackage{algorithm}
\usepackage{algpseudocode}
\usepackage{subcaption}
\usepackage{multirow}
\usepackage{graphicx}
\usepackage{array}
\usepackage{changepage}
\usepackage{caption}

\newcommand{\methodname}{Cost and Reward Infused Metric Elicitation}
\newtheorem{assumption}{Assumption}
\newtheorem{proposition}{Proposition}

\title{\methodname}

\author{%
  Chethan Bhateja \\
  Department of Computer Science\\
  Stanford University\\
  Stanford, CA 94305 \\
  \texttt{chethanb@stanford.edu} \\
  \And
  Joseph O'Brien \\
  Department of Computer Science \\
  Stanford University \\
  Stanford, CA 94305 \\
  \texttt{jobrien3@stanford.edu} \\
  \And
  Afnaan Hashmi \\
  Department of Computer Science \\
  Stanford University \\
  Stanford, CA 94305 \\
  \texttt{afnaan@stanford.edu} \\
  \And
  Eva Prakash \\
  Department of Computer Science \\
  Stanford University \\
  Stanford, CA 94305 \\
  \texttt{eprakash@stanford.edu} \\
}

\begin{document}

\maketitle

\begin{abstract}
In machine learning, metric elicitation refers to the selection of performance metrics that best reflect an individual's implicit preferences for a given application. Currently, metric elicitation methods only consider metrics that depend on the accuracy values encoded within a given model's confusion matrix. However, focusing solely on confusion matrices does not account for other model feasibility considerations such as varied monetary costs or latencies. In our work, we build upon the multiclass metric elicitation framework of \citet{multiclass}, extrapolating their proposed Diagonal Linear Performance Metric Elicitation (DLPME) algorithm to account for additional bounded costs and rewards. Our experimental results with synthetic data demonstrate our approach's ability to quickly converge to the true metric. Our code can be found at \href{https://github.com/chethus/metric}{https://github.com/chethus/metric}.
\end{abstract}

\section{Introduction}
The selection of performance metrics in machine learning is crucial to ensuring they align with real-world application needs and effectively capture the trade-offs and priorities of an intended use case. For example, in the context of a cost-sensitive predictive categorical model for cancer diagnosis \citep{Yang2014-mf}, failing to place sufficient importance on false negatives can lead to missed or delayed disease detection, whereas false positives can lead to unnecessary treatment. Both cases can have severe health consequences for patients. While a doctor can make an informed decision on patient care based on years of training, the challenge arises in determining how to choose a model that best reflects the doctor's preferences. In other words, we would like to translate the knowledge of a user (i.e. the doctor) about the trade-offs between models into a numerical form. This numerical form is known as the performance metric \citep{10.1145/3375627.3375862,10.1145/2983323.2983356}. 

Motivated by such scenarios, metric elicitation is a framework to identify the underlying performance metric of a practitioner in a given context \citep{hiranandani2019performancemetricelicitationpairwise}. This task is accomplished by relying on user feedback with the goal of collecting as few queries as possible due to the high cost of acquiring feedback from individuals \citep{hiranandani2019performancemetricelicitationpairwise}. The general framework for metric elicitation was first coined by \citet{hiranandani2019performancemetricelicitationpairwise}, utilizing the confusion matrices of classifiers to derive a metric function in a binary classification setting. Due to the inaccuracies that generally result from asking people to give their absolute preferences \citep{qian2013active}, the authors proposed to present the oracle (i.e. the user) with pairwise preferences between confusion matrices \citep{hiranandani2019performancemetricelicitationpairwise}. Traditionally, the metric function takes a classifier's confusion matrix as input and outputs a measure of how much the user prefers this classifier \citep{hiranandani2019performancemetricelicitationpairwise}. \citet{multiclass} expanded upon the binary classification setting, proposing methods to derive the metric for a classification task with an arbitrary number of classes. 

However, the confusion matrix omits other important attributes of a model that may matter to users. For example, work has pointed towards a potential trade-off between accuracy and fairness \citep{https://doi.org/10.1002/int.22354,accuracyTradeoff,menon2018cost}. In turn, \citet{hiranandani2020fairperformancemetricelicitation} adapted the multiclass metric elicitation framework by introducing terms into the metric to penalize large differences in accuracy between different groups. Still, this work did not account for additional trade-offs that could exist beyond fairness. 

Other work has centered on analyzing the trade-off between accuracy and other model-related criteria. \citet{costAccuracyTradeoff} explored the relationship between accuracy and model training time by varying the number of iterations and amount of data used during training. The authors also investigated how execution time related to accuracy in an ensemble setting by altering the number of models used in the ensemble \citep{costAccuracyTradeoff}. \citet{10.1145/3578356.3592578} highlighted the trade-off between latency, accuracy, and cost-efficiency for inference serving systems. The authors framed the scenario as a multi-objective constraint satisfaction problem formulated using integer linear programming. While the authors propose an objective function to maximize with weights determined by user preference, they did not offer a method to identify this function \citep{10.1145/3578356.3592578}.

Thus, we see that relying solely on models' confusion matrices can exclude other relevant factors. Practical applications often involve trade-offs that go beyond accuracy. For example, a less accurate model might be chosen due to lower monetary costs, or a faster model might be favored when latency constraints are critical. These additional factors can be essential for determining the most appropriate classifier in real-world scenarios. Thus, running our algorithm could enable corporations to choose models with other desirable attributes, such as serving clients efficiently or being environmentally friendly.

In this work, we explore the question of how costs and rewards for a specific classifier can be incorporated into metric elicitation. To this end, we propose a novel algorithm for the elicitation of metrics which include bounded classifier-specific costs and rewards.
\begin{figure}
   \centering
   \includegraphics[scale=0.4]{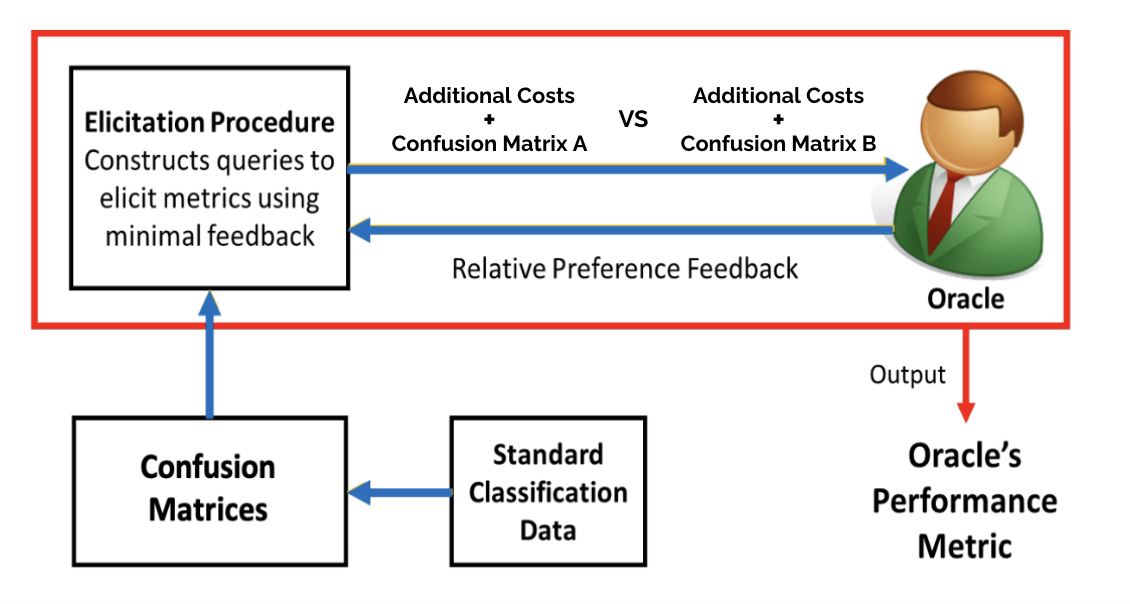} 
   \caption{ Cost and reward infused metric elicitation framework adapted from \citet{hiranandani2019performancemetricelicitationpairwise}.}
   \label{fig:metric_elicitation_image}
\end{figure}

\section{Preliminaries}
We first define relevant variables following the notation described in \cite{hiranandani2019performancemetricelicitationpairwise}. Metric elicitation can be performed on binary classifiers by making use of confusion matrices to recover the true performance metric $\phi$ of an oracle.
In this setting, $X$ is defined as the features of a single data point and $Y$ is defined as the corresponding binary label (0 = negative class, 1 = positive class). $\eta(x) = P(Y=1|X=x)$ represents the conditional probability of the positive class ($Y=1$), and $\zeta = P(Y=1)$ represents the unconditional probability of the positive class. A classifier $h$ maps the featurized input $X$ to one of the two classes with the goal of accurately approximating the relationship between $X$ and $Y$ through maximizing the relevant performance metrics derived from the confusion matrix. The confusion matrix contains rates for true positives (TP), true negatives (TN), false positives (FP), and false negatives (FN) defined as follows
\begin{itemize}
    \item $TP(h, P) = P(Y = 1, h = 1)$
    \item $FP(h, P) = P(Y = 0, h = 1)$
    \item $FN(h, P) = P(Y = 1, h = 0)$
    \item $TN(h, P) = P(Y = 0, h = 0)$
\end{itemize}
The confusion matrix $C$ can be reduced to a two-dimensional space of just $TP$ and $TN$, as $FN(h, P)$ can be expressed as $\zeta - TP(h, P)$ and $FP(h, P)$ can be expressed as $1 - \zeta - TN(h, P)$.


The true performance metric $\phi$ for metric elicitation is estimated by querying an oracle $\Omega$ with pairwise comparisons of the confusion matrices of different classifiers, which is represented as the following:
$$
\Gamma(h, h') = \Omega(C, C') = 1[\phi(C) > \phi(C')].
$$
The error between the true metric $\phi$ and resulting estimated metric $\hat{\phi}$ is written as $\|\phi - \hat{\phi}\|$.

As described by \citet{multiclass}, the multiclass setting builds upon the binary setting by instead classifying a featurized datapoint $X$ into one of $k$ classes $(Y \in [k])$. The conditional probability is modified from the binary setting to be $\eta_i(x) = P(Y=i| X=x)$ to represent the likelihood of class $i$ for the input, and $\zeta_i = P(Y=i)$ now represents the unconditional probability of class $i$. The classifier $h$ maps the featurized input $x$ to a probability for each class $i$ in $k$. The confusion matrix becomes a $k$ x $k$ matrix where diagonal entries (predicted class = true class) indicate correct classifications and off-diagonal entries (predicted class $\neq$ true class) indicate incorrect classifications. The probability of the predicted class being $j$ and the true class being $i$ is represented as follows in the confusion matrix:
$$
C_{ij}(h, P) = P(Y = i, h = j)
$$
Similar to the binary setting, the simplification of $C_{ii}$ can be represented as follows:

$$\zeta_i - \sum_{j \neq i} C_{ij}(h)$$
Diagonal elements capture the probability of correct predictions for each class (class accuracy) while off-diagonal elements capture the probability of errors.

One of the types of performance metrics used in multiclass classification settings is the Diagonal Linear Performance Metric (DLPM), which only considers class accuracies when evaluating a classifier. The weight matrix $a \in \mathbb{R}^k$ with $\|a\|_1 = 1$ to ensure scale invariance is used to weight the individual importances of each class in $k$. A DLPM metric $\psi$ in the family of DPLM metrics $\phi_\text{DLPM}$ is formulaically defined as the following, with $\mathbf{d} = \text{diag}(C(h, P))$ being the diagonal elements of $C(h, P)$:

$$\psi_\text{DLPM}(\mathbf{d}) = \langle \mathbf{a}, \mathbf{d} \rangle = \sum_{i=1}^k \mathbf{a}_i \mathbf{d}_i$$


To apply this multiclass framework to our setting where a classifier $h$ can have bounded costs and rewards, we introduce a vector of reward terms $\mathbf{r}(h)$ and cost terms $\mathbf{c}(h)$. Each reward term $r_i(h)$ has a bounded range $[0, A_i]$ for all classifiers and similarly each cost term $c_i(h)$ has a bounded range $[-B_i, 0]$. Then for our cost and reward infused diagonal linear performance metric, we denote the weights for the accuracies in $\mathbf{d}(h)$ by $\mathbf{a^d}$, the weights for the rewards in $\textbf{r}(h)$ by $\mathbf{a^r}$, and the weights for costs $\textbf{c}(h)$ by $\mathbf{a^c}$. With this notation, a linear performance metric over accuracies, costs, and rewards is defined as
\begin{equation*}
\psi(\mathbf{d}, \mathbf{r}, \mathbf{c}) = \langle \mathbf{a^d}, \mathbf{d} \rangle + \langle \mathbf{a^c}, \mathbf{c} \rangle + \langle \mathbf{a^r}, \mathbf{r} \rangle
\end{equation*}
As a shorthand, we also denote the concatenation of $(\mathbf{a^d}, \mathbf{a^r}, \mathbf{a^c})$ as $\mathbf{a}$.

\section{Geometry of the Space of Cost and Reward Infused Classifiers}
We first analyze the query space of classifiers for cost and reward infused metric elicitation, which augments the space of multiclass confusion matrices in multiclass metric elicitation \citep{multiclass} with bounded costs and rewards. As in multiclass metric elicitation \citep{multiclass}, we make the following regularity assumption on the input data distribution of the classifier:
\begin{assumption}[Trade-off Between Accuracies]
Assume that $\forall i, j \in [k], g_{ij}(r)=P \left [\frac{\eta_i(X)}{\eta_j(X)} \geq r \right ]$ are continuous and strictly decreasing for $r \in [0, \infty)$.
\end{assumption}
Intuitively, this assumption means that there is always a trade-off between the accuracies of different classes. Therefore changing the relative weights of the accuracies will change the preferences over classifiers. Given this assumption, the multiclass metric elicitation paper finds the following:
\begin{proposition}[Confusion Space]
The space of diagonal confusions $\mathcal{D}$ is strictly convex, closed, and contained in the box $[0, \zeta_1] \times [0, \zeta_2] \times \dots \times [0, \zeta_k]$. The vertices of the space of diagonal confusions are given by $\zeta_i \mathbf{e}_i \forall i \in [k]$, where $\mathbf{e}_i$ denotes the $i$th elementary vector.
\end{proposition}
\begin{figure}
   \centering
   \includegraphics[scale=0.3]{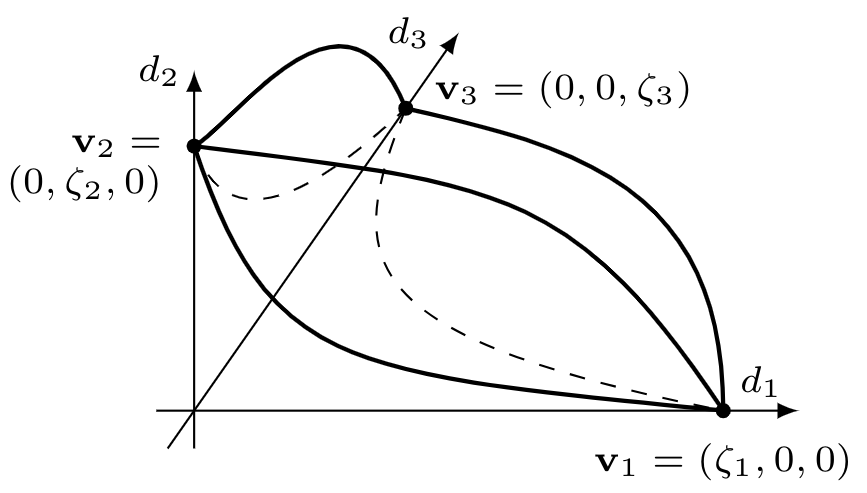} 
   \caption{Confusion Space of Classifiers for $k=3$ \citep{multiclass}}
   \label{fig:confusion_space}
\end{figure}
We include a helpful visualization of the confusion space from \citet{multiclass} in Figure \ref{fig:confusion_space}. Note that each vertex $\zeta_i \mathbf{e}_i$ of the space of confusions is achieved by the trivial classifier $\forall x, h(\mathbf{x})=i$ which predicts only class $i$ for all data points.
Extending this space to include a bounded cost $c_i$ with maximum magnitude $B_i$ adds a dimension with range $[-B_i, 0]$. Similarly, extending this to a beneficial feature of a classifier $r_i$ with maximum magnitude $A_i$ adds a dimension with range $[0, A_i]$. \\
Key to our algorithm is the assumption of an implicit trade-off between costs/rewards and accuracies in our data. Without this assumption, we cannot obtain queries that trade-off between these features, and it is impossible to determine how one should be exchanged for the other.
\begin{assumption}[Trade-off Between Costs/Rewards and Accuracy]
\label{asm:cost_accuracy}
For any reward $\mathbf{r}_i$, the Pareto frontier $f_{ij}(t) = \max_{\substack{h \in \mathcal{H} \\ \mathbf{d}_j(h) \geq t}} \mathbf{r}_i(h)$ is strictly decreasing. Similarly, for any cost $\mathbf{c}_i$, the Pareto frontier $f_{ij}(t) = \max_{\substack{h \in \mathcal{H} \\ \mathbf{d}_j(h) \geq t}} \mathbf{c}_i(h)$ is strictly increasing.
\end{assumption}
In our later experiments with synthetic data, we choose an $f_{ij}$ that takes on all possible slopes in $[0, \infty]$, guaranteeing that we can approximate the true metric for any set of weights.
\section{Method}
Our algorithm will consider different sets of hypothesis weights, determine the optimal classifier for each set of weights, query the oracle for comparisons of these classifiers, and use the query outputs to narrow the search space of weights.
\subsection{Choosing Hypothesis Weights}
Akin to DLPM, our metric only depends on the ratios between the weights, corresponding to the direction of the normalized weight vector. We follow the strategy in \citet{multiclass} of making comparisons that change two classifier statistics at a time between statistics $\mathbf{d^1}, \mathbf{c^1}, \mathbf{r^1}$ and $\mathbf{d^2}, \mathbf{c^2}, \mathbf{r^2}$ because this depends only on the ratio between weights. \\

As a concrete example, the output of a deterministic oracle when changing only accuracies $\mathbf{d}_i$ and $\mathbf{d}_j$ for $j \neq i$ is determined by the difference in metrics
\begin{align*}
\psi(\mathbf{d^1}, \mathbf{c^1}, \mathbf{r^1}) - \psi(\mathbf{d^2}, \mathbf{c^2}, \mathbf{r^2}) =& \langle \mathbf{a^d}, \mathbf{d^1} \rangle + \langle \mathbf{a^c}, \mathbf{c^1} \rangle + \langle \mathbf{a^r}, \mathbf{r^1} \rangle - \langle \mathbf{a^d}, \mathbf{d^2} \rangle - \langle \mathbf{a^c}, \mathbf{c^2} \rangle - \langle \mathbf{a^r}, \mathbf{r^2} \rangle \\
=& \mathbf{a^d_i} (\mathbf{d^1}_i - \mathbf{d^2}_i) - \mathbf{a^d_j} (\mathbf{d^1}_j - \mathbf{d^2}_j) \\
\propto & \frac{\mathbf{a^d}_i}{\mathbf{a^d}_j} (\mathbf{d^1}_i - \mathbf{d^2}_i) - (\mathbf{d^1}_j - \mathbf{d^2}_j)
\end{align*}
which we can see only depends on the ratio $\frac{\mathbf{a^d}_i}{\mathbf{a^d}_j}$. To take advantage of this observation, we can sample different weight ratios for each classifier attribute relative to the class 1 accuracy with weights for all other statistics set to 0. Then we find the optimal classifier for each of these samples and query the oracle with the corresponding metrics, enabling us to binary search on the true weight ratio.
\subsection{Determining the Optimal Classifier}
When searching for a class accuracy, cost, or reward weight relative to the first weight (class 1 accuracy), our framework has two cases that we must consider differently. The first case is that we are searching for the weight of an accuracy and the second is that we are searching for the weight of a cost or reward.
\subsubsection{Optimal Weighted Accuracy Classifier}
As noted in \citet{multiclass}, the optimal classifier that predicts only two classes for a weighted accuracy metric is given by the Restricted Bayes Optimal (RBO) classifier.
\begin{proposition}[RBO Classifier]
For a weighted accuracy metric $\psi \in \varphi_{DLPM}$, the optimal classifier that predicts only distinct classes $k_1, k_2 \in [k]$ is given by
\[
\bar{h}_{k_1, k_2}(x)=
\begin{cases}
k_1 & a_{k_1} \eta_{k_1}(x) \geq a_{k_2} \eta_{k_2}(x) \\
k_2 & else
\end{cases}
\]
\end{proposition}
When our hypothesis weights are only on two accuracies, we know that only these two corresponding classes will be predicted and the optimal classifier will be the RBO classifier for the corresponding two classes.
\subsubsection{Cost-Accuracy Classifier}
\label{subsec:optimalCost}
The optimal classifier for a metric that weights only class 1 accuracy and a single cost/reward can be easily determined when the function for the Pareto frontier given by $f_{ij}(t)$ in Assumption \ref{asm:cost_accuracy} is known. To find this classifier, we can solve for the point where the slope of the Pareto frontier matches the slope of the metric's level curves which is determined by the ratio of the metric weights.\\
\begin{figure}
   \centering
   \includegraphics[scale=0.5]{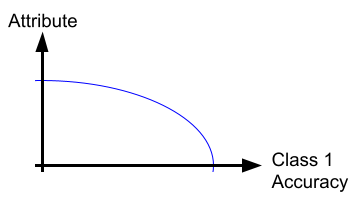} 
   \caption{We choose the Pareto frontier to be a quarter ellipse to cover all metric slopes. Note that "Attribute" above can either correspond to a cost or a reward. If "Attribute" is a reward the vertical y axis is positive, and if "Attribute" is a cost then the vertical y axis is negative.}
   \label{fig:tradeoff}
\end{figure} \\
It is possible this point does not exist if the Pareto frontier does not contain all possible slopes. When generating synthetic data, we choose a quarter ellipse as the Pareto frontier as shown in Figure \ref{fig:tradeoff}, covering all possible weight ratios (slopes) in $[0, \infty]$ between the two attributes. For the particular case of the ellipse with a maximum reward of $A$, we derive that the angle of the optimal classifier on the ellipse is given by
\begin{equation*}
\theta = \tan^{-1} \left ( \frac{\eta_1 m}{A (1 - m)} \right )
\end{equation*}
This angle $\theta$ determines a point on the ellipse, giving us the optimal classifier's cost/reward and accuracy values. We implement the reduction in accuracy according to the Pareto frontier by introducing noise through a binomial probability $p_\text{wrong}$, a chance of predicting an impossible class for a data point $\mathbf{x}$ regardless of the class probabilities.
\subsection{Algorithm}
Our detailed pseudocode is shown in Algorithm \ref{alg:cdlpm}. For each possible attribute, our algorithm samples five possible weight ratios compared to the class 1 accuracy weight. It constructs an RBO classifier for accuracy attributes or a cost-accuracy/reward-accuracy optimal classifier for non-accuracy attributes. Then we use the outputs of the queries to narrow the search space of the optimal weight ratio in a binary search based approach similar to prior work \citep{hiranandani2019performancemetricelicitationpairwise}, \citep{multiclass}. \\

Given $n$ possible classifier attributes, including $k$ class accuracies and other cost and reward attributes, our algorithm runs an independent binary search on each attribute. Therefore, the runtime to achieve error $\epsilon$ is $O(n \log(\frac{1}{\epsilon}))$
\begin{algorithm}
\caption{\methodname}\label{alg:cdlpm}
\textbf{Input:} $\epsilon > 0$, oracle $\Gamma$, attribute\_spaces (lower and upper bounds)
\begin{algorithmic}
\Procedure{elicitation}{oracle, $\epsilon$}
\State $\hat{\mathbf{a}} = \mathbf{0}$
\State $\mathbf{\hat{a}^d}_1 = 1$
\For{$i \leftarrow 2,\dim(\hat{\mathbf{a}})$}
\State $a, b = 0, 1$
\While{$b - a > \epsilon$}
\State $c, d, e = \frac{3a + b}{4}, \frac{a + b}{2}, \frac{a + 3b}{4}$
\For{$x \in (a, c, d, e, b)$}
\State{// Check if $\hat{\mathbf{a}}_i$ is an accuracy attribute}
\If{$i < \dim(\mathbf{\hat{a}^d})$}
\State $h_x = \textsc{CONSTRUCT\_RBO\_CLASSIFIER}(x, i)$
\Else
\State $h_x = \textsc{CONSTRUCT\_COST\_ACCURACY\_CLASSIFIER}(x, i, \text{attribute\_spaces})$
\EndIf
\EndFor
\State $q_{ac}, q_{cd}, q_{de}, q_{eb}=\Gamma(h_a, h_c), \Gamma(h_c, h_d), \Gamma(h_d, h_e), \Gamma(h_e, h_b)$
\If {$q_{ac}$ or $q_{cd}$}
\State b = d
\ElsIf{$q_{de}$}
\State a = c
\State b = e
\Else
\State a = d
\EndIf
\EndWhile
\State mid = (a + b)/2
\State $\hat{\mathbf{a}}_i = (1 - \text{mid}) / \text{mid}$
\EndFor
\State \Return $\text{normalize}(\hat{\mathbf{a}})$
\EndProcedure
\Statex
\Procedure{Construct\_RBO\_Classifier}{m, i}
\State $\hat{\mathbf{a}} = m \mathbf{e}_1 + (1 - m) \mathbf{e_i}$
\State \Return Classifier(weights=$\hat{\mathbf{a}}$, $p_\text{wrong}=0$)
\EndProcedure
\Statex
\Procedure{Construct\_Cost\_Accuracy\_Classifier}{m, i, attribute\_spaces}
\State $\hat{\mathbf{a}} = m\mathbf{e}_1, \mathbf{r} = \mathbf{0}, \mathbf{c} = \mathbf{0}$
\State lower, upper = attribute\_spaces[i].lower, attribute\_spaces[i].upper
\State $\theta = \tan^{-1} \left (\frac{\eta_1 m}{(\text{upper} - \text{lower})(1-m)} \right )$
\If{$i \leq \dim(\mathbf{\hat{a}^d}) + \dim(\mathbf{\hat{a}^r})$}
\State ind $= i - \dim(\mathbf{\hat{a}^d})$
\State $\mathbf{r}_{\text{ind}} = \text{lower} + (\text{upper} - \text{lower}) \cos(\theta)$
\Else
\State ind $= i - \dim(\mathbf{\hat{a}^d}) - \dim(\mathbf{\hat{a}^r})$
\State $\mathbf{c}_\text{ind} = \text{lower} + (\text{upper} - \text{lower}) \cos(\theta)$
\EndIf
\State \Return Classifier(weights=$\hat{\mathbf{a}}, \mathbf{r}, \mathbf{c}, p_\text{wrong}=1 - \sin(\theta)$)
\EndProcedure
\end{algorithmic}
\end{algorithm}
\section{Experiments and Analysis}

Similar to \citet{hiranandani2019performancemetricelicitationpairwise} and \citet{multiclass}, we utilize synthetic data to verify our approach. To represent our input, we generate a set $\{\mathbf{x^i}\}^n_{i=1}$ of 10-dimensional vectors, where the entries of a given $\mathbf{x_i}$ are sampled from $\mathcal{N}(0,1)$. For our experiments, we set $n=10,000,000$. For our output, we assume that there are $k$ classes. To derive $\eta_i({\mathbf{x}}$), we first generate a $k \times 10$ weight matrix with entries sampled from $\mathbb{U}[-1.5,1.5]$. We then multiply this weight matrix by each of our n input vectors and take the softmax result to generate the desired probabilities. We specify the lower bound $-B_i$ for cost attributes and the upper bound $A_i$ for reward attributes. These bounds determine the shape of the quarter ellipse described earlier in section \ref{subsec:optimalCost}. 

\captionsetup[table]{aboveskip=5 pt}
\renewcommand{\arraystretch}{1.5}
\begin{table}[h!]
\centering
\scalebox{0.785}{
\begin{tabular}{|c|c|c|c|}
\hline
\multicolumn{2}{|c|}{Classes $k$ = 2, Reward/Cost Attributes = 2} & \multicolumn{2}{c|}{Classes $k$ = 3, Reward/Cost Attributes = 3} \\
\hline
$\psi$* = $\mathbf{a}$* & $\hat{\psi}$ = $\hat{\mathbf{a}}$ & $\psi$* = $\mathbf{a}$* & $\hat{\psi}$ = $\hat{\mathbf{a}}$ \\
\hline
(0.10, 0.05, 0.05, 0.80) & (0.10, 0.05, 0.05, 0.80)  & (0.12, 0.08, 0.07, 0.32, 0.19, 0.22) & (0.12, 0.08, 0.07, 0.32, 0.19, 0.22) \\
(0.32, 0.17, 0.28, 0.23) & (0.32, 0.17, 0.28, 0.23) & (0.12, 0.08, 0.07, 0.32, 0.19, 0.22) & (0.12, 0.08, 0.07, 0.32, 0.19, 0.22) \\
\hline
\end{tabular}
}
\caption{Cost and reward infused metric elicitation with $\epsilon = 0.001$ for synthetic data. Table structure inspired by \citep{multiclass}.}
\label{tab:psiTable}
\end{table}
We first took an approach akin to that of \citet{multiclass} to establish the validity of our method, eliciting underlying metrics with varied numbers of classes and additional attributes. We performed experiments with the following setup reflected in Table \ref{tab:psiTable} above:

\begin{itemize}
    \item $k=2$ classes, 1 reward attribute with range [0,5], 1 cost attribute with range [-0.3,0]
    \item $k=2$ classes, 2 cost attributes with ranges [-15,0] and [-18,0]
    \item $k=3$ classes, 2 cost attributes with ranges [-8,0] and [-20,0], 1 reward attribute with range [0,15.5]
    \item $k=3$ classes, 1 cost attribute with range [-0.5,0], 2 reward attributes with ranges [0,20] and [0,30]
\end{itemize}

From the table, we can recognize that our elicited weights $\hat{\mathbf{a}}$ well reflect the true weights $\mathbf{a}$*. We note that in the table we round our derived metric weights to two decimal places due to spacing constraints, and that the derived weights slightly diverge from the true metric when we include higher precision in the resulting values. Nonetheless, this divergence is slim and the resulting values still closely match the true weights. In both cases with $k=2$ and two attributes, there were 120 total queries. In both cases with with $k=3$ and three attributes, there were 200 total queries.

\begin{figure}[htbp]
  \centering
  \begin{minipage}[b]{0.45\textwidth}  
    \centering
    \raisebox{6mm}{\includegraphics[scale=0.4]{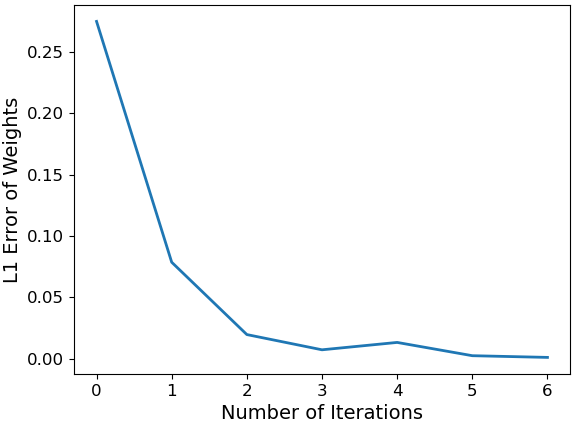}
    \label{fig:image1}}
  \end{minipage}%
  \hspace{0.0cm}  
  \begin{minipage}[b]{0.45\textwidth}  
    \centering
    \includegraphics[scale=0.6]{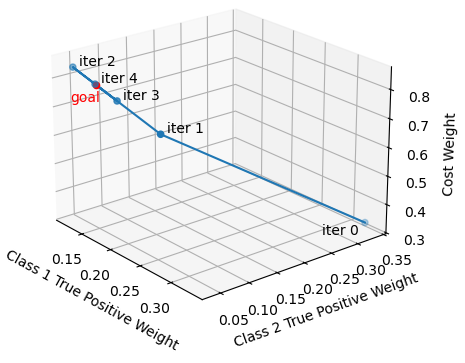}
    \label{fig:image2}
  \end{minipage}
  \caption{Plot showing the L1 error with each passing iteration (left). 3D representation of the algorithm learning with two classes and a single additional cost (right).}
  \label{fig:sidebyside}
\end{figure}

To better illustrate how our algorithm tries to derive the true weights, we plotted both the L1 error \\||$\hat{\mathbf{a}}$ - $\mathbf{a}$*||${_1}$ as the algorithm progressed and a 3D representation of how $\hat{\mathbf{a}}$ moves towards the true metric weights over time as shown in Figure \ref{fig:sidebyside}. In this experiment, instead of binary searching until our weights reach a desired precision $\epsilon$, we instead set the number of iterations of binary search. For the L1 error plot, we use the same weights as the first entry in Table \ref{tab:psiTable} for $k=2$ with 2 attributes. For the 3D representation, we also utilize $k=2$ for our classes with true weights of 0.15 and 0.5 and a single cost with true weight 0.8.

In both representations, we can see the logarithmic runtime of the algorithm at play. The L1 error plot shows that the error decreases logarithmically at each step until convergence, as is expected due to binary search. When the estimated metric weights near the true values the L1 error becomes non-monotonic, but we ascribe these fluctuations to floating point precision error. The 3D plot similarly shows logarithmic convergence in all 3 dimensions of the weights. The plot also provides a nice illustration of the binary search based algorithm at play. From iterations 0 through 1, we can deduce that the lower bound of our weight estimate must be updated as the weights increase closer to the goal. By iteration 2, the true weights are overestimated, and instead the upper bound is decreased. On iteration 3, we again underestimate the true weights as the lower bound is too low. By iteration 4, we see that our estimated weights have essentially converged to the true metric weights.

\section{Future Work}

In our current implementation, we only adapt the DLPM approach of multiclass metric elicitation \citet{multiclass} to account for additional features beyond the confusion matrix. This approach does not allow the metric to distinguish between different types of misclassifications (off-diagonal values of the confusion matrix). Thus, our method could be adapted to instead build off the more general LPM to account for these false classifications in the metric. In a similar vein, future work could also explore adapting Linear-Fractional Performance Metric Elicitation (LFPME) to account for linear-fractional metrics such as F1 score \citep{multiclass}.

As described in our experiments section, we currently only test our approach using synthetic data with the trade-off between a given class accuracy and cost/reward attribute dictated by the quarter ellipse Pareto frontier seen in Figure \ref{fig:tradeoff}. To further evaluate the robustness of our method, this experimentation could be extended to real-world datasets. Beyond requiring access to additional model statistics to describe the additional attributes of note such as monetary cost and inference speed, a way to define the true trade-off between model accuracies and these additional cost/reward attributes would be required. If we are only allowed to make queries with respect to classifiers that actually exist, our algorithm with synthetic data wouldn't be applicable. In such cases with an unknown Pareto frontier, we could also run a bisection method to determine the optimum if we assume that the frontier is strictly decreasing.

Furthermore, if the utilities of these extra features are non-linear, then our method would not be at all applicable, as this linear nature is what allows us to make pairwise comparisons to derive the metric weights. An example of this type of feature that is neither strictly good nor bad could be the length of a response provided by a large language model (LLM). Initially, an increase in response length could correspond to an increase in utility due to an increase in clarity and user understanding, but after reaching an inflection point increasing the length could decrease utility due to the response becoming overly complicated and drawn out. To address this non-monotonic behavior, an approach that is akin to gradient descent may be of interest. In this vein, we can see how small changes in the value of the non-linear feature affect our resulting estimate of the metric and continuously move in the direction that maximizes our understanding of user preference. However, we leave further exploration of this idea for future work.

Lastly, another avenue for potential exploration would be adapting the method to account for group-based metrics. Currently, we assume that costs are a fixed attribute of the classifier but it is possible costs could vary on different subsets of the data, just like accuracies in fair performance metric elicitation \citep{hiranandani2020fairperformancemetricelicitation}. For example, consider a setting where different data points are from different countries and costs vary by country. This scenario demonstrates how combining cost and reward infused metric elicitation with the underlying ideas of fair performance metric elicitation of \citet{hiranandani2020fairperformancemetricelicitation} could be a beneficial expansion of our work.

\section{Ethics and Society Review Statement}
One risk of our metric elicitation tool is that our algorithm learns the preferred metrics and costs of an individual practitioner and an individual practitioner only. We learn and choose the model that best matches a given user’s trade-off of accuracies, costs, and rewards. What may be ideal for one group of practitioners is not necessarily aligned with the preferences of all users, such as their firm or organization.

As an example, we can imagine a work context where our algorithm is used to help determine which models to employ. A specific user who has an incentive to finish his given deliverables rapidly would prioritize speed over monetary cost and would reveal those preferences to our algorithm. This could be misaligned with the preferences of his firm, which could have to pay the costs of a more costly but quicker model. Similarly, a practitioner who neglects the computational power and monetary cost required for a model would cause our algorithm to select a computationally expensive model that presents an environmental cost that may be undesirable for their broader society. In a medical context, a doctor might have a higher preference for model speed, as they would like treat as many patients as possible. However, an individual patient could have a higher preference for maximizing accuracy due to the potential serious consequences of an incorrect diagnosis.

To mitigate such scenarios, we recommend a mandatory educational workshop or tutorial that must be completed by any potential users of our algorithm. In such a workshop, we would try to impart an understanding of the existence and preferences of other stakeholders who would be affected by the usage of our metric elicitation algorithm. Informing users that any costs they will not personally incur will be incurred somewhere else would help align individual preferences with the preferences of all other potential stakeholders.

Another risk we identified was the potential reputation damage that could arise from data leaks. Metric elicitation is often used in a medical context to improve diagnosis. Medical practitioners could potentially apply this algorithm to determine how they are willing to trade off costs/rewards with accuracy and use this metric to select their preferred classifier. Metric elicitation in a medical context can reveal the inherent operating trade-offs of accuracy and model costs. If this information were released to the public, it could risk causing a deluge of negative press, as the perceived valuation of false negative or false positive predictions alongside other cost/reward attributes by a medical practitioner or system could be controversial.

Any deployment of our algorithm into real-world contexts must have strict, up-to-date security features to minimize the chance of a data breach causing furor. In addition, our model should encrypt learned values for various parameters as much as possible so that even authorized users of the system cannot go into the code and reveal the exact learned cost trade-off.

An additional risk could arise when practitioners are unaware or unsure of their own preferences. People often struggle to accurately identify what they truly want or what is best in a given situation. A practitioner may believe they understand their preferences regarding trade-offs, yet their actual decisions could be misaligned with the trade-offs they intend to prioritize in a specific context.

To address this, we recommend slowing down the deployment of our algorithm. Before running our algorithm, we will first include a required survey to for users to fill out. This survey would include questions pertaining to the importance to be ascribed to rewards and costs such as speed, accuracy, monetary cost, etc. The survey would not affect our algorithm directly, but would serve to get the practitioner to think critically about their own values. This period of reflection will help to align the practitioner’s choices with their actual preferences.

Moreover, yet another risk that arises would be ill-intentioned practitioners using our algorithm to obfuscate responsibility. Some practitioners may have personal incentives to prefer models with rewards, costs, and accuracies that most benefit themselves even if they know this preference negatively impacts others. For example, a practitioner could choose a model by prioritizing speed and low monetary cost, even if accuracy takes a hit. To deflect blame from negatively impacted individuals, the practitioner might point to the seemingly black box nature of the elicitation algorithm, even though the algorithm actually is based on their true metric.

To address this potential ethical concern, we would mandate a clause that must be agreed to before the use of our algorithm. This clause would stipulate that an explanation of what our algorithm accomplishes and how it is used in the specific work pipeline must be provided to any relevant parties. This will ensure these impacted individuals are aware that the algorithm learns the preferences of the practitioner and will help mitigate the possibility of our algorithm being used in poor taste.

\section{Conclusion}
In this work, we studied current literature in the realm of metric elicitation and proposed a new algorithm, cost and reward infused metric elicitation, that takes into account attributes beyond confusion matrices. Our proposed algorithm successfully elicited the true metric when tested on synthesized data. Future work can expand upon our approach by testing our algorithm on real data and exploring ways to address non-linear costs and rewards.

\bibliographystyle{acl_natbib}
\bibliography{citations}


\end{document}